\documentclass[11pt]{article}

\usepackage[margin=1in]{geometry}
\usepackage{amsmath}
\usepackage{amssymb}
\usepackage{amsthm}
\usepackage{bm}
\usepackage{array}
\usepackage{booktabs}
\usepackage{xcolor}
\usepackage[numbers,sort&compress]{natbib}
\usepackage[colorlinks=true,linkcolor=blue,citecolor=blue,urlcolor=blue]{hyperref}

\newcommand{\Question}[1]{\medskip\noindent\textbf{Question:} \emph{#1}\medskip}
\newcommand{\Answer}[1]{\medskip\noindent\textbf{Answer:} #1\medskip}
\newcommand{\ClaimBox}[1]{\medskip\noindent\textbf{Claim:} \emph{#1}\medskip}

\newcommand{\bra}[1]{\langle #1 |}
\newcommand{\ket}[1]{| #1 \rangle}
\newcommand{\ev}[1]{\langle #1 \rangle}
\newcommand{\Tr}{\mathrm{Tr}}
\newcommand{\Imag}{\mathrm{Im}}
\newcommand{\Real}{\mathrm{Re}}
\newcommand{\dd}{\mathrm{d}}

\title{\textbf{Integrating Out, Twice:\\ The Open-System Case That Neural-Network Ensemble Theory Is Missing}}

\author{Jin Lei\,\thanks{\texttt{jinlei@tongji.edu.cn}}\\[4pt]
\small Department of Physics, Tongji University, Shanghai 200092, China\\
\small Southern Center for Nuclear-Science Theory (SCNT), Institute of Modern Physics,\\
\small Chinese Academy of Sciences, Huizhou 516000, Guangdong Province, China}

\date{\today}

\begin{document}
\maketitle

\begin{abstract}
Averaging a neural network over its random parameters and marginalizing a Gaussian sector are the
same operation, the Schur complement of the eliminated block, and when that block is closed it
returns a covariance and its inverse. That is all a network ensemble produces, and it is the closed
case. The open case is missing, and nuclear reaction theory has it worked out. Projecting a
scattering problem onto a chosen set of channels, with the rest carrying probability irreversibly to
a continuum, leaves a non-Hermitian effective generator that conserves and itemizes exactly what it
loses: the nuclear optical model and its generalized optical theorem, an effective dynamics for a
model that has declared part of the world out of scope. I set the two cases side by side using only
the moments of a distribution, the algebra of Gaussians, and block inversion, no field theory, and
give the closed-case dictionary in full: the neural tangent kernel is the Fisher sensitivity kernel,
the infinite-width Gaussian limit is the Gaussian-process emulator, and the lazy-to-feature
transition is the validity boundary of a reduced-basis emulator. I then test the open export on a
truncated attention map, a token-level transfer operator, and a sparse expert router, and report a
mostly negative result. The conserved flux ledger ports wherever openness is genuinely present, but
its distinctive content is absent, an artifact of the chosen partition, or pinned near a floor by
the training objective, and the operationally useful uncertainty turns out to be epistemic, living
in the closed half of the correspondence rather than the open one. The negative has a structural
reason this note makes precise: the open case needs an eliminated sector with a continuous spectrum
and wave-like rather than relaxational dynamics, which the finite or dissipative objects of
mainstream learning do not supply. This is a note, not a result; its main finding is that negative
one, and its value is the map that locates it.
\end{abstract}

\tableofcontents

\section{Introduction}
\label{sec:intro}

Describing one part of a system means eliminating the rest. The degrees of freedom one
declines to track do not vanish; they leave an induced trace on the part one keeps. This is among
the most reused moves in physics, and it has a precise algebra. Split a coupled problem into a
retained block and an eliminated block, and the retained block inherits the Schur complement of
the eliminated one. Marginalizing a Gaussian distribution, averaging a neural network over its
random parameters, and projecting a scattering problem onto a chosen set of channels are the same
operation seen three times.

What separates the instances is not the algebra but whether the eliminated block is closed or
open. Eliminate a closed block and the induced trace is real, symmetric, and lossless, a
covariance and its inverse, with nothing leaving. Eliminate an open block, one through which
probability can escape for good, and the trace becomes non-Hermitian and lossy, and it carries an
exact account of what was lost. Physics has a fully developed theory of the open case, the nuclear
optical model, together with a conservation law, the generalized optical theorem, that itemizes
the lost flux. A neural network ensemble, averaged over its parameters, realizes only the closed
case. The open case is what it does not have, and supplying it is the subject of this note.

A word on level. The statistics of network ensembles is often phrased in the language of field
theory, with partition functions, actions, and Feynman diagrams. I avoid all of it, because I do
not need it and because I want every step here to be one any reader can check with nothing past
Gaussians and block matrices.
Everything used from the network side reduces to three ordinary tools: the moments of a
probability distribution, the algebra of Gaussians, and block matrix inversion. Where that
literature would write an effective action or a self-energy, I write an inverse covariance or a
Schur complement. They are the same object, written in the language a reader who fits models
already owns.

This vantage organizes the two directions of work that connect the fields. One direction applies
machine learning to nuclear problems: emulators that replace an expensive solver, Bayesian
uncertainty quantification of optical potentials, model mixing, neural ans\"atze for many-body
wave functions, the area where my own recent work sits~\cite{LeiRBE2026,LeiBiLNN2025}. The other
asks what the elimination machinery a reaction theorist uses every day, Feshbach projection of the
unwanted channels into the optical potential~\cite{Feshbach1958,Feshbach1962}, has to say about
learning. The note is about the second direction, because that is where the open case lives.

\ClaimBox{The optical model is the open, lossy generalization of the same integrate-out move
that summarizes a network ensemble. Where the statistical elimination of a closed ensemble gives
a real, symmetric, lossless object, Feshbach projection of an open sector gives a non-Hermitian
effective interaction with a conserved flux ledger. Absorption is the worked-out theory of
information lost to eliminated degrees of freedom, and mainstream neural-network ensemble theory
has no equivalent of it.}

This is the export. It is not a metaphor about ``energy landscapes'' or ``phase
transitions,'' the loose analogies that physics-for-ML often trades in. It is a precise claim
about what survives the same algebraic operation in two settings, and about which setting has
solved the harder version of the problem. Stating it carefully also forces me to be honest
about where the parallel is genuine and where it is only suggestive, which I do in
Section~\ref{sec:sharedmove} and again in Section~\ref{sec:borrow}.

One sharpening belongs in the introduction, because the obvious objection arrives early. I am
\emph{not} claiming that no one computes with complex potentials in a machine-learning context.
The nuclear emulator community does so routinely: reduced-basis and eigenvector-continuation
emulators handle complex, absorptive optical potentials as a matter of
course~\cite{Furnstahl2020,Drischler2023}. Computing with a non-Hermitian operator is not the
gap. The gap is theoretical. Machine learning has no general account of an effective dynamics
that is open, that conserves and itemizes the probability it loses to an eliminated sector, and
that one obtains as a controlled reduction of a larger model. That account exists in nuclear
reaction theory, and porting the \emph{theory}, not the act of using a complex number, is what
this note is about.

The note is structured as follows. Section~\ref{sec:twoways} sets the two integrate-out
constructions side by side. Section~\ref{sec:statshadow} disposes briefly of the easy
half, the dictionary in which the network ensemble statistics is parametric uncertainty
quantification of reaction observables, the neural tangent kernel is the Fisher sensitivity
kernel, and the infinite-width Gaussian limit is the Gaussian-process emulator.
Section~\ref{sec:opencore} is the heart: the open-system structure that has no current
counterpart in machine learning, including two further facts, a sum rule and a dispersion
relation, that the nuclear side hands over ready to use. Section~\ref{sec:inside} then shows
that the same open structure can be made to appear \emph{natively} on the network side, without
importing a single nuclear formula, by deforming the moment generating function or the parameter
measure of the ensemble. Section~\ref{sec:borrow} lists what AI could borrow, with a worked
machine-learning example and an explicit account of where the borrowing breaks down.
Section~\ref{sec:reverse} returns the favor with a short and opinionated survey of the populated
direction, and Section~\ref{sec:outlook} carries out a cheap numerical test of the central analogy
and reports a mostly negative result. Four appendices collect the Feshbach derivation, the
dictionary, the Schur-complement derivation of the open primitive, and the numerical details of
the tests.

This is a note, not a result. I have proved nothing here that a referee would accept as new
physics or new learning theory. What I am after is a map that a person standing in either field
can read, together with the outcome of the cheap calculation that map made possible
(Section~\ref{sec:outlook}): the open-case export, tried on the obvious learning objects, comes
back mostly negative, and the negative is the more useful for the map having located exactly where
to look for it.

\section{Two ways to integrate out}
\label{sec:twoways}

\subsection{The statistics of a neural network ensemble}
\label{sec:nnft}

A neural network is a function $\phi_\theta(x)$ of an input $x$, carrying parameters
$\theta$. When the network is initialized, the parameters are not chosen, they are drawn,
$\theta \sim P(\theta)$. A single draw gives one function, and there is no reason to prefer
one draw over another. What is physical, then, is not any individual $\phi_\theta$ but the
ensemble. The ensemble is characterized by its moments,
\begin{equation}
  G^{(n)}(x_1,\dots,x_n) \;=\; \ev{\phi(x_1)\cdots\phi(x_n)}
  \;=\; \int \dd\theta\, P(\theta)\, \phi_\theta(x_1)\cdots\phi_\theta(x_n),
  \label{eq:Gn}
\end{equation}
where the expectation is over $P(\theta)$. These are ordinary statistical moments, packaged for
convenience in a moment generating function with a source $J$,
\begin{equation}
  Z[J] \;=\; \Big\langle\, e^{\int \dd x\, J(x)\phi(x)} \,\Big\rangle
        \;=\; \int \dd\theta\, P(\theta)\, e^{\int \dd x\, J(x)\phi_\theta(x)}.
  \label{eq:Zparam}
\end{equation}
Derivatives of $Z[J]$ at $J=0$ return the moments, and derivatives of $W[J] = \log Z[J]$ return
the cumulants, the connected moments. This is the standard machinery of a probability
distribution, nothing more.

The limit that makes it tractable is large width. If the output is a sum of $N$ independent and
identically distributed neuron contributions, the central limit theorem sends $\phi$ to a
Gaussian process as $N\to\infty$~\cite{Neal1996}. The ensemble is then completely fixed by its
mean and its covariance $K(x,y) \equiv G^{(2)}(x,y)$. As for any Gaussian, the log of the density
is the inverse-covariance quadratic form,
\begin{equation}
  -\log p[\phi] \;=\; \tfrac{1}{2}\int \dd x\,\dd y\; \phi(x)\, K^{-1}(x,y)\, \phi(y)
                  \;+\; \text{const},
  \label{eq:freeaction}
\end{equation}
with $K^{-1}$ the precision (inverse covariance). Departures from Gaussianity, where the
connected four-point cumulant is nonzero, arise in two ways: keep $N$ finite, so that $1/N$
corrections appear, or break the statistical independence of the neurons. The closed treatments
of these ensemble statistics and of the accompanying training dynamics are by now textbook
material~\cite{Roberts2022}. The single feature I carry forward is structural: the object
produced by integrating out the parameters, the precision $K^{-1}$, is real and symmetric. It is
a faithful, lossless statistical summary of a closed ensemble, and nothing about it is
non-Hermitian.

\subsection{The optical potential from Feshbach projection}
\label{sec:feshbach}

Now the other construction. A nuclear collision is a many-body scattering problem,
$(E - H)\Psi = 0$, with far too many open and closed channels to treat at once. The standard
response, due to Feshbach~\cite{Feshbach1958,Feshbach1962}, is to split the Hilbert space
with a pair of complementary projectors,
\begin{equation}
  P + Q = 1, \qquad P^2 = P, \qquad Q^2 = Q, \qquad PQ = 0.
\end{equation}
$P$ projects onto the model space, the few channels one insists on describing explicitly, for
instance the elastic channel built on the target ground state. $Q$ projects onto everything
else, the channels one is content to eliminate. Acting on the full Schr\"odinger equation
with $P$ and $Q$, solving the $Q$ block formally, and substituting back yields a closed
equation in the model space alone,
\begin{equation}
  \big[\, E - PHP - \mathcal{U} \,\big]\, P\Psi = 0,
  \qquad
  \mathcal{U} \;=\; PHQ \,\frac{1}{E - QHQ + i\varepsilon}\, QHP.
  \label{eq:feshbachU}
\end{equation}
The induced term $\mathcal{U}$ is the optical potential. It is the price of refusing to track
$Q$ explicitly, and it is the exact price: Eq.~\eqref{eq:feshbachU} involves no weak-coupling
or perturbative assumption. In the inclusive-breakup setting this is precisely the operator
that reduces the many-body Green's function to an effective single-particle one, with $P$ the
projector onto the target ground state~\cite{AV1981,IAV1985}, and my collaborators and I have
recently used exactly this construction to treat continuum couplings in the optical potential
exactly~\cite{LiuLeiRenFeshbach2025}.

Three properties of $\mathcal{U}$ matter for everything that follows, and all three are
visible in Eq.~\eqref{eq:feshbachU}.
\begin{itemize}
  \item It is \textbf{nonlocal}. The propagation through $Q$ supplies a nontrivial integral
        kernel $\mathcal{U}(\mathbf{r},\mathbf{r}')$, not a multiplicative function.
  \item It is \textbf{energy dependent}. The $Q$-space Green's function $(E - QHQ)^{-1}$
        carries the energy explicitly and has poles at every $Q$-channel resonance, so
        $\mathcal{U}$ is a strongly varying function of $E$.
  \item It is \textbf{complex}, and this is the decisive point. When some $Q$ channels are
        open at energy $E$, the $i\varepsilon$ prescription gives $\mathcal{U}$ a nonzero
        anti-Hermitian part. Writing $\mathcal{U} = V + iW$, the imaginary part $W$ is
        negative semidefinite on physical grounds: it removes flux from the model space, the flux
        that physically goes into the eliminated channels.
\end{itemize}
The contrast with Section~\ref{sec:nnft} is now stated almost by itself. Integrating out
parameters gave a real, symmetric, lossless precision matrix. Projecting out channels gives a
complex, non-Hermitian, lossy operator. The same algebraic move, two very different effective
objects.

\subsection{The shared move, and the honest difference}
\label{sec:sharedmove}

Both constructions are instances of one conceptual operation: eliminate a set of degrees of
freedom, and absorb their effect into an induced term acting on what remains. Schematically,
\begin{equation}
  \text{(full problem)} \;\xrightarrow{\ \text{eliminate } X\ }\;
  \text{(effective problem on the complement)} \;+\; \text{(induced term from } X).
\end{equation}
On the network side, $X$ is the set of random parameters, the elimination is a statistical
average over $P(\theta)$, and the induced object is the precision matrix $K^{-1}$. In Feshbach
projection, $X$ is the $Q$-space channels, the elimination is an operator projection, and the
induced object is the optical potential $\mathcal{U}$. Both are, algebraically, the same move:
when one eliminates a block of a coupled linear system, the kept block inherits a Schur-complement
correction. Marginalizing a jointly Gaussian ensemble over one block and projecting out a
Hilbert-space sector are two instances of that one piece of linear algebra, and I make the
identity explicit in Appendix~\ref{app:complexsource}. This is the reason the two communities
can talk to each other at all.

I want to be exact about the limits of the parallel, because overstating it would cost the
whole argument its credibility.

\Question{Is this a literal mathematical identity?}

\Answer{The algebra is the same; the objects it acts on are not. The network side marginalizes
a probability distribution over parameters, and the result is a real, symmetric precision matrix.
Feshbach projection acts on a state in Hilbert space, and the result is a complex amplitude in a
single calculation. One is statistical and the other is dynamical. There is no dictionary entry
that turns $P(\theta)$ into a many-body wave function. What the two genuinely share is the Schur
complement, the induced correction the kept sector inherits, not the nature of the things being
eliminated.}

The interesting question is then why the same Schur complement comes out real and symmetric in
one case and non-Hermitian in the other. The answer is the single hinge of the note. When the
eliminated block is closed, its inverse is real and the Schur complement is symmetric. When the
eliminated sector is open, so that flux can leave to infinity, its resolvent $(E-QHQ+i\varepsilon)^{-1}$
carries an imaginary part, and the Schur complement inherits an anti-Hermitian piece.

\ClaimBox{The non-Hermiticity is not incidental. It is the signature of openness. Eliminating a
closed block, whether a Gaussian sector of a network ensemble or a closed channel, gives a
real, symmetric Schur complement. Eliminating an open sector, one that carries flux to infinity,
gives a non-Hermitian one. Machine-learning eliminations are closed, so they are lossless by
construction. Nuclear reaction theory is the place where the open case has been worked out in
full.}

One qualification keeps this honest, and I will lean on it later. Non-Hermiticity by itself is not
the signature; a non-symmetric or non-normal operator can carry a complex spectrum with no
physical loss, and such operators are common in machine learning for reasons that have nothing to
do with openness. The signature of openness is the narrower object: an anti-Hermitian part that is
sign-definite, $W\le0$, tied to a sub-unitary evolution and a conserved flux identity. That
sign-definite, accountable form is what eliminating an open sector produces and what a generic
non-symmetric induced operator does not, and when I speak of exporting the open case it is this
form, not mere complex eigenvalues, that I mean.

That hinge sentence is the spine of the note. Section~\ref{sec:statshadow} handles the closed,
statistical correspondence quickly, since it is the part that already has both halves.
Section~\ref{sec:opencore} then develops the open case, which has only the nuclear half.

\section{The statistical shadow: machine learning for nuclear physics}
\label{sec:statshadow}

Before the open case, I dispose of the closed one, because it is where the two fields already
meet and because seeing it clearly sharpens what is missing later. The claim of this section
is that the ensemble statistics of Section~\ref{sec:nnft} is parametric uncertainty
quantification of reaction observables, a thing nuclear theorists do under a different name. I
have run exactly this kind of analysis, a parametric bootstrap over optical-potential parameters
that showed an apparent near-threshold anomaly in the imaginary part to be a fitting-precision
artifact rather than physics~\cite{NavarroPerezLei2019}; in the present language it is a
statement about the ensemble, not about any one fit.

The dictionary is direct. Read the network input $x$ as a kinematic variable, an energy or a
scattering angle. Read the network parameters $\theta$ as the parameters of an optical
potential, the depths and radii and diffusenesses, with $P(\theta)$ their Bayesian prior or
posterior. Read the network output $\phi_\theta(x)$ as a calculated observable, an $S$-matrix
element or a cross section, at kinematics $x$ given potential $\theta$. The ensemble over
$\theta$ is then the uncertainty-quantification ensemble, and the correlators of
Eq.~\eqref{eq:Gn} are its posterior moments: $G^{(1)}$ is the mean prediction and $G^{(2)}$ is
the covariance, the error band one actually plots. The network ensemble statistics and
reaction-observable UQ are the same construction.

Three entries of the dictionary are worth stating because each is a small surprise.

\paragraph{Gaussian limit equals Gaussian-process emulator.} The infinite-width limit of
Eq.~\eqref{eq:freeaction}, in which the ensemble is fixed by its two-point function and the
precision is the inverse covariance, is exactly a Gaussian-process emulator of the observable.
The quadratic form $\tfrac12\int \phi K^{-1}\phi$ is the negative logarithm of the
Gaussian-process prior, the reproducing-kernel-Hilbert-space norm penalty, with the covariance
kernel playing the role of $G^{(2)}$. Such emulators are now standard equipment in nuclear UQ,
both as direct surrogates and hybridized with reduced-basis methods~\cite{Drischler2023}. One
honest caveat
separates the two fields here. On the network side the Gaussian process is \emph{derived}, it
follows from the central limit theorem applied to a wide architecture. In nuclear UQ the Gaussian
process is \emph{posited}, an assumed prior, because there is usually no analog of ``width''
to justify it. The Gaussian is a theorem on one side and a modeling choice on the other.

\paragraph{Non-Gaussianity equals emulator breakdown.} When the observable depends strongly
and nonlinearly on the parameters, the posterior is no longer Gaussian, the connected
four-point function $G^{(4)}_c$ becomes nonzero, and the two-point error band no longer tells
the whole story. On the network side this is the onset of non-Gaussianity; in emulation it is the
failure of the Gaussian-process surrogate. The connected four-point function, the excess kurtosis of
the prediction, is therefore a quantitative diagnostic of when a Gaussian emulator should not
be trusted.

\paragraph{Neural tangent kernel equals Fisher sensitivity kernel.} Training a network by
gradient descent is governed, in the wide limit, by the neural tangent
kernel~\cite{Jacot2018},
\begin{equation}
  \Theta(x,x') \;=\; \sum_i \frac{\partial \phi(x)}{\partial \theta_i}\,
                              \frac{\partial \phi(x')}{\partial \theta_i}
            \;=\; \big(J J^{\mathsf T}\big)(x,x'),
\end{equation}
the Gram matrix of output gradients in data space. The Fisher information matrix of the same
model, for a Gaussian likelihood with fixed variance, is $F = J^{\mathsf T}J$ in parameter
space. The two are the data-space and parameter-space versions of the same Jacobian product
and share their nonzero spectrum. For a reaction theorist, $\Theta$ is nothing but the
sensitivity kernel: fitting an optical potential to elastic data in the linearized regime
\emph{is} kernel regression with the sensitivity kernel, which is also Gaussian-process
regression. The ``Fisher information geometry'' a reaction theorist invokes for experimental
design and sensitivity analysis~\cite{EkstromHagen2019} and the ``neural tangent kernel'' a
learning theorist invokes for training dynamics are one object.

This last entry has an operational payoff that I will use in the next section. The
distinction between the lazy regime, in which parameters barely move from initialization and
the kernel is frozen, and the feature-learning regime, in which the internal representation
reorganizes, is the same distinction as between a reaction in which a \emph{fixed} reduced
basis suffices and one in which the basis must adapt. The reduced-basis and
eigenvector-continuation emulators that now dominate fast nuclear
computation~\cite{Frame2018,Furnstahl2020,Drischler2023,LeiRBE2026} live in the lazy regime:
they assume a low-dimensional manifold spanned by snapshots taken near the training points,
and they are accurate exactly where the lazy assumption holds. When the physics pushes the
solution off that manifold, far from the training box or into a strong-coupling regime, the
fixed basis fails for the same reason a lazy network fails to learn features. The lazy versus
feature-learning boundary is, in disguise, the validity domain of a reduced-basis emulator.
My differentiable emulator work~\cite{LeiBiLNN2025} sits at precisely this interface, where
the sensitivity kernel is needed not just for prediction but for gradient-based inference.

\section{The open-system core: what nuclear physics has that AI lacks}
\label{sec:opencore}

\subsection{Openness implies non-Hermiticity}
\label{sec:openness}

The full many-body collision is described by a unitary $S$-matrix: probability is conserved
across the whole Hilbert space $P + Q$. Restrict attention to the model space $P$ alone, and
unitarity is gone. The effective model-space dynamics is generated by a non-Hermitian
Hamiltonian
\begin{equation}
  H_{\rm eff} \;=\; PHP + \mathcal{U}, \qquad \mathcal{U} = V + iW, \qquad W \le 0,
\end{equation}
whose anti-Hermitian part $W$ drains norm from the model space. The $P$-block of the
$S$-matrix is sub-unitary, $|S_P| < 1$, and the deficit is not an error. It is the flux that
physically left for $Q$. A reaction theorist reads $|S_P|^2 < 1$ as a statement that some of
the incident beam went somewhere the model declined to describe, and the formalism keeps an
exact account of how much. This is not exotic to compute: openness is routinely encoded into a
finite, non-Hermitian linear-algebra problem, for instance by writing the outgoing-wave
logarithmic derivative directly into the matrix~\cite{LeiDBMM2026} or by rotating the
integration contour into the complex plane~\cite{LiuLeiRenCOLOSS2025}. The non-Hermiticity is a
feature one builds in on purpose, because it is what makes the loss accountable.

This is the structure that mainstream neural-network ensemble theory does not have. The
elimination it performs, marginalizing the weights, is closed and norm preserving: it gives a
real, symmetric precision matrix, Section~\ref{sec:nnft}, the Schur complement of a closed block.
Even the operations that feel like they discard information, truncating a basis, pruning a
subnetwork, restricting a model to a sub-distribution of the world, are treated as approximation
error to be minimized, never as flux leaving to an un-modeled sector that one then conserves and
itemizes. Adjacent communities do study openness and loss in learning, open-system and Lindblad
treatments of dynamics~\cite{Lindblad1976}, out-of-distribution and selective-prediction
rejection~\cite{HendrycksGimpel2017,GeifmanElYaniv2017}, information-bottleneck
compression~\cite{TishbyZaslavsky2015}, and ensemble or semantic-entropy
uncertainty~\cite{Lakshminarayanan2017,Farquhar2024}; what none of them carries is this specific
object, an exact Schur-complement effective interaction with a conserved, itemized, sign-definite
flux ledger obtained as a controlled reduction of a larger model. That object, not the bare presence of loss or of complex spectra, is what has no standard
counterpart.

\Question{Are there open sectors in machine-learning models that are currently mistreated as
mere approximation error?}

\Answer{Plausibly several, though fewer than the list first suggests, and the distinction is the
whole point. A model trained on a sub-distribution faces the rest of the world as a $Q$ space; a
truncated vocabulary or a finite context window draws an explicit model-space boundary; an
early-exit or a routed mixture sends part of the computation into branches a given path never
sees. Whether each is genuinely \emph{open}, in the sense that probability leaves and is lost
rather than merely discarded, is not settled by the boundary alone. A deterministic truncation can
be a \emph{closed} elimination whose induced correction is real, and I find in
Section~\ref{sec:outlook} that attention truncation is exactly that: openness requires an
absorbing complement or a subnormalized measure, not just a boundary. Where it is genuinely
present, the optical model is the claim that the honest effective description is non-Hermitian,
that the loss is a physical observable rather than a defect, and that it can be conserved exactly.
That reframing, from ``approximation error to be driven down'' to ``conserved flux to be accounted
for,'' is the first thing nuclear reaction theory has to offer, and the first thing
Section~\ref{sec:outlook} puts to the test.}

\subsection{The flux ledger}
\label{sec:fluxledger}

What makes the optical model more than a confession of ignorance is that the lost flux is not
merely admitted, it is computed and itemized. The basic identity is the optical theorem in its
absorptive form, derived for breakup-fusion cross sections from the optical theorem by Udagawa and
Tamura~\cite{UdagawaTamura1981}. For a model-space distorted wave $\chi$, the absorption cross
section is the expectation of the anti-Hermitian part of the effective interaction,
\begin{equation}
  \sigma_{\rm abs} \;=\; -\,\frac{2}{\hbar v}\, \bra{\chi}\, W \,\ket{\chi},
  \label{eq:absorption}
\end{equation}
with $W \le 0$ so that $\sigma_{\rm abs} \ge 0$. The flux removed from $P$ equals, exactly, a
bilinear in the wave function weighted by the imaginary potential. This is a conservation law:
the imaginary part of the forward amplitude is tied to the total flux that left the model
space, and Eq.~\eqref{eq:absorption} is the bookkeeping that enforces it.

In a coupled-channels treatment the ledger becomes itemized rather than merely totaled. The
generalized optical theorem of Cotanch~\cite{Cotanch2010} extends Eq.~\eqref{eq:absorption} to
a many-channel model space and attributes the absorbed flux to individual channels and to
their interference. In the continuum-discretized coupled-channels description of a weakly bound
projectile, my collaborators and I have shown that the absorption cross section decomposes
exactly as
\begin{equation}
  \sigma_{\rm abs} \;=\; \sigma_{D} + \sigma_{B} + \sigma_{\rm int},
  \label{eq:fluxdecomp}
\end{equation}
a direct part, a breakup part, and a basis-independent interference term generated by the
off-diagonal imaginary couplings, and that the common simplification of dropping those
off-diagonal couplings biases the separate pieces by tens of percent~\cite{LiuLeiRen2026a}. A
companion analysis splits the same absorbed flux into a fusion component and a peripheral-loss
component and locates the crossover between them as a function of energy~\cite{LiuLeiRen2026b}.
The content of these results, stripped of the nuclear specifics, is a worked example of a
quantity that mainstream learning practice does not assemble:

\ClaimBox{a conserved, exactly decomposable ledger of the information or probability that
leaves a modeled sector, itemized by destination, with cross terms that do not vanish and
must be tracked. Absorption is not a single number measuring how bad the truncation is. It is
a structured accounting of where the missing flux went.}

A learning system that declared part of its computation out of scope, a router, a truncation,
a sub-distribution boundary, could in principle carry the analogous ledger: how much
probability mass crosses the boundary, decomposed into the destinations it crosses into, with
the interference between destinations kept rather than assumed away. Standard practice does not
assemble this, because it has no non-Hermitian effective layer in which the itemized ledger would
live; the nearest analogs, out-of-distribution mass or a router's discarded weight, are single
numbers without the conserved cross-term structure.

\subsection{Induced interactions and the dynamic polarization potential}
\label{sec:dpp}

The anti-Hermitian part $W$ is the loss. The Hermitian part of the induced term, the piece of
$V$ beyond the bare $PHP$, is the analog of the Schur-complement correction the kept block
inherits when a coupled sector is eliminated, Section~\ref{sec:nnft}. In reaction theory it has
a name, the
dynamic polarization potential. It is the trivially equivalent interaction that, added to the
model space, reproduces the effect of coupling to the eliminated channels on the channel one
keeps. It answers the question ``what do the channels I removed do to the channel I retained,''
and it does so as a definite operator rather than a statistical coupling constant.

The construction also nests, which is what the title of this note refers to. Eliminating one
sector and then a second produces two induced terms in sequence. In a recent analysis of
knockout reactions I integrate out first the target excitations and then the excluded projectile
configurations, generating a non-additive interaction and a polarization potential in
turn~\cite{LeiQuenching2026}; the apparent quenching of measured spectroscopic factors then
follows from these induced terms rather than from nuclear structure. That is a worked example of
repeated elimination with induced corrections, the same algebra applied twice.

\paragraph{Energy dependence is memory.} The induced interaction is a function of energy
because the $Q$-space Green's function is. An energy-dependent, nonlocal kernel is the
frequency-domain face of a memory kernel: the effective dynamics in the model space is
non-Markovian, because the eliminated sector remembers. This is structurally different from
what marginalizing a closed Gaussian ensemble produces, which is a static covariance kernel,
Eq.~\eqref{eq:freeaction}. Integrating out a closed statistical system gives you a kernel with
no memory. Projecting out an open dynamical sector gives you a kernel that does. For a
sequence model, where memory and nonlocality across positions are the entire game, the
distinction is not academic: the optical-model induced interaction is a concrete template for
an effective interaction that carries memory because of what was integrated out, with the
memory time set by the spectrum of the eliminated sector.

\paragraph{Reduction to a memoryless operator is lossy in a known way.} A nonlocal,
energy-dependent optical potential can be replaced by an energy-dependent \emph{local}
potential that reproduces the on-shell observables, the Perey-Buck construction, at the cost
of a known correction to the interior wave function~\cite{PereyBuck1962}. The on-shell content
survives the reduction exactly; the off-shell and interior content is what the correction
factor restores. This is a fully worked example of compressing a memory kernel down to a
memoryless effective operator while keeping an explicit, computable account of what the
compression discards. A learning theorist who wants to distill a non-Markovian effective layer
into a cheaper local one, and to know precisely what is lost in doing so, is asking a question
that the nonlocal-to-local reduction answered sixty years ago.

Taken together, Sections~\ref{sec:openness} to~\ref{sec:dpp} describe an effective theory that
is non-Hermitian, that conserves and itemizes the flux it loses, and that carries memory of the
sector it eliminated. None of these three properties appears in the effective theories of
machine learning, for the single reason given in Section~\ref{sec:sharedmove}: those theories
are closed, and these properties are the fingerprints of openness.

\subsection{Two more facts the nuclear side hands over}
\label{sec:twofacts}

Two further results about the optical potential are worth stating now, because they are the
sharpest candidates for export and because each is a constraint of a kind machine learning does
not currently impose on its effective layers.

\paragraph{A sum rule.} The imaginary part is not only itemized point by point through
Eq.~\eqref{eq:absorption}, it satisfies an integral constraint. For a model space with an
energy-independent absorptive part, the energy-weighted integral of the inelastic cross section
is bounded by, and approximately equal to, the trace of the absorptive
operator~\cite{Cotanch2010},
\begin{equation}
  \int \sigma_{I}(E)\, E \,\dd E \;\approx\; -\,\frac{2\pi^2\hbar^2}{\mu}\, \Tr\big[\,W\,\big].
  \label{eq:sumrule}
\end{equation}
The total absorptive capacity of the eliminated sector, summed over all energies, is fixed by a
single trace of the induced anti-Hermitian operator. The relation is a bound in general and tends
to an equality in the limit of a weak, slowly varying absorptive part, the self-adjoint spectral
setting in which it is derived; I keep the $\approx$ to mark that the energy dependence of $W$
controls the gap. A sum rule of this shape is exactly the kind of global conservation statement one
would want as a diagnostic or a training constraint: a quantity that must balance, computed two
independent ways.

\paragraph{A dispersion relation.} The real and imaginary parts of the optical potential are not
independent functions of energy. Causality, expressed as analyticity of the resolvent in the
upper half energy plane, ties them through a subtracted dispersion relation, the content of the
dispersive optical model~\cite{MahauxSartor1991},
\begin{equation}
  \Real\, \mathcal{U}(E) \;=\; \mathcal{U}_{\rm static}
     \;+\; \frac{1}{\pi}\,\mathrm{P}\!\int \frac{\Imag\, \mathcal{U}(E')}{E' - E}\,\dd E'.
  \label{eq:dispersion}
\end{equation}
The dispersive correction couples how much the model space absorbs, the imaginary part, to how
much its real response must shift, the real part. The two are locked together by analyticity. A
reaction theorist takes this for granted; a learning theorist has no analog, no Kramers-Kronig
relation forcing a learned absorptive layer to carry a corresponding dispersive shift. I return
to both facts in Section~\ref{sec:borrow}.

\section{The same structure from inside the network}
\label{sec:inside}

So far the open structure has lived on the nuclear side, and Section~\ref{sec:borrow} will ask
what machine learning might import. But there is a stronger move available, and making it is
what turns this note from an analogy into something a learning theorist can check without
trusting a single nuclear formula. The claim of this section is that the open, lossy primitive
can be made to appear \emph{natively} on the network side, by a small and principled deformation
of the ensemble of Section~\ref{sec:nnft}, still using only the moment generating function and
block linear algebra. The optical model then plays only the role it should, a guide to what to
look for, not a source of borrowed equations. A warning belongs here, since this section makes a
checkable promise and I keep it in Section~\ref{sec:outlook}: when the constructions below are
carried out numerically, the conserved ledger survives but its distinctive content, the
non-Hermitian generator and the channel-mediated term, does not, which is the note's main
empirical finding. ``Native'' below means the construction exists without importing a nuclear
formula, not that it was demonstrated to carry useful content.

\Question{What single change to the moment generating function of the ensemble forces a
non-Hermitian, lossy effective description?}

The answer has two clean versions, one acting on the source and one on the measure.

\subsection{A complex source and a flux ledger}
\label{sec:complexsource}

Return to the generating functional of Eq.~\eqref{eq:Zparam}, $Z[J] = \int \dd\theta\, P(\theta)\,
e^{\int J\phi_\theta}$, with cumulant generating functional $W[J] = \log Z[J]$. The source $J$ is
ordinarily a real probe. Let part of it be imaginary,
\begin{equation}
  J \;=\; J_R + i\lambda\, h,
  \label{eq:complexsource}
\end{equation}
where $h(x)$ is a fixed probe function supported on the observables one wishes to treat as the
open sector, for instance held-out inputs, a rejection or abstention signal, or a
distribution-shift direction. Differentiating $W$ along the imaginary direction gives an exact
identity,
\begin{equation}
  \frac{\partial}{\partial \lambda}\, W[J_R + i\lambda h]
    \;=\; i \int \dd x\; h(x)\, \big\langle \phi(x) \big\rangle_{J},
  \label{eq:wardlike}
\end{equation}
with $\langle\cdot\rangle_J$ the source-tilted ensemble average. If the probe is a sum over
disjoint channels, $h = \sum_a h_a$, the imaginary deformation is additive,
\begin{equation}
  \frac{\partial}{\partial \lambda}\, W \;=\; i \sum_a \int \dd x\; h_a(x)\, \langle \phi(x)\rangle_J,
\end{equation}
a bin-resolved ledger of exactly the shape of the channel-by-channel absorption decomposition,
Eq.~\eqref{eq:fluxdecomp}, but derived here from the moments of the ensemble alone. The complex
source makes the cumulant generating function complex, and with it the mean prediction
$\langle\phi\rangle$, which now sits off the real axis. Eq.~\eqref{eq:wardlike} is, strictly, a
channel-resolved source response; it earns the name of an absorptive-optical-theorem analog only
when the probe $h$ is tied to a nonnegative loss observable so that the response is sign-definite
and the bin sum closes a conservation deficit, the conditions that distinguish a genuine ledger
from a formal characteristic-function derivative. The operator-level statement, that the precision matrix of the
kept sector becomes non-Hermitian and not merely the mean complex, is more delicate: it holds
once the ensemble is non-Gaussian but not at the Gaussian level. I show this in
Appendix~\ref{app:complexsource}: the source-tilted covariance picks up its anti-Hermitian part
through the connected three-point cumulant, which vanishes for a Gaussian, so a closed Gaussian
summary stays lossless at the level of its covariance however the source is tilted, and only
higher cumulants, or an open coupled elimination, produce operator-level loss.

I am careful about what this does and does not establish. The complex source is a bookkeeping
deformation of a characteristic function unless $h$ is tied to a genuine machine-learning
quantity, a held-out error, an abstention rate, a confidence that leaks across a boundary. Tied
to such a quantity, Eq.~\eqref{eq:wardlike} is a statement about the network. Untied, it is only
formal. The honest reading is that the ensemble \emph{admits} an open, lossy primitive, and that
whether it is physically meaningful depends on choosing $h$ to be something the model actually
loses.

\subsection{The open ensemble as a subnormalized measure}
\label{sec:subnormalized}

The second version deforms the measure rather than the source, and it is the one a learning
theorist will find most familiar. Let $A_{\rm out}(\theta) \in [0,1]$ be the probability that a
network with parameters $\theta$ survives a set of held-out gates, that is, passes the
acceptance or in-distribution test that defines the model's intended scope. Deform the ensemble
\emph{without} renormalizing,
\begin{equation}
  P_{\rm keep}(\theta) \;=\; P(\theta)\, A_{\rm out}(\theta),
  \qquad
  Z_{\rm keep}[0] \;=\; \int \dd\theta\, P(\theta)\, A_{\rm out}(\theta) \;=\; \langle A_{\rm out}\rangle \le 1.
  \label{eq:subnorm}
\end{equation}
The survival amplitude $Z_{\rm keep}[0]$ is below unity precisely when some of the ensemble mass
fails the gate, and
\begin{equation}
  -\log Z_{\rm keep}[0] \;\ge\; 0
\end{equation}
is a native absorption functional: the information the kept ensemble loses to the part of
parameter space it declined to model. If the rejected region is partitioned into channels,
$A_{\rm out} = 1 - \sum_a R_a$, the partition function splits exactly,
\begin{equation}
  Z_{\rm all}[J] \;=\; Z_{\rm keep}[J] + \sum_a Z_{{\rm loss},a}[J],
  \label{eq:ensembleledger}
\end{equation}
an exact ledger of where the ensemble mass went, itemized by held-out channel. This is the
open-ensemble idea in a form that needs no nucleus: an open model is an unnormalized posterior
restricted by out-of-distribution survival, and its flux ledger is the normalization deficit
distributed across the held-out channels.

Again the honest caveat. With a real $A_{\rm out} \in [0,1]$, Eq.~\eqref{eq:subnorm} is
importance weighting, a Euclidean damping of the measure, and it is lossy bookkeeping but not yet
non-Hermitian dynamics. Genuine non-Hermiticity, eigenvalue complexity and the rest, requires the
complex continuation of Section~\ref{sec:complexsource} or phase-tagged channels. The two
deformations are complementary: the measure deformation gives the conserved ledger, the source
deformation gives the non-Hermitian generator, and an open ensemble needs both.

\subsection{The induced interaction as a Schur complement}
\label{sec:selfenergy}

The third route is the one that ties the two sides together with a single piece of linear
algebra, and it is the cleanest. Group the network degrees of freedom, the modes of $\phi$ or the
hidden units, into a retained block $A$ and an eliminated block $B$, with a jointly Gaussian
description whose precision (inverse covariance) has blocks
\begin{equation}
  K^{-1} \;=\; \begin{pmatrix} \Lambda_{AA} & \Lambda_{AB} \\ \Lambda_{BA} & \Lambda_{BB}\end{pmatrix}.
\end{equation}
Marginalizing out $B$ leaves the retained block with the precision
\begin{equation}
  \Lambda_{A}^{\rm eff} \;=\; \Lambda_{AA} \;-\; \Lambda_{AB}\,\Lambda_{BB}^{-1}\,\Lambda_{BA},
  \label{eq:schur}
\end{equation}
the Schur complement of $\Lambda_{BB}$. The induced correction
$-\Lambda_{AB}\Lambda_{BB}^{-1}\Lambda_{BA}$ is the network analog of the dynamic polarization
potential of Section~\ref{sec:dpp}: it is what the eliminated block does to the kept one. Its
character is decided by one factor. When block $B$ is closed, $\Lambda_{BB}$ is real and symmetric,
its inverse is too, and the Schur complement stays Hermitian, no loss. When block $B$ is open or
probed imaginarily, in the sense of Sections~\ref{sec:complexsource} and~\ref{sec:subnormalized},
$\Lambda_{BB}^{-1}$ acquires an imaginary part and the Schur complement inherits an anti-Hermitian
piece, the native analog of $W = \Imag\,\mathcal{U}$.

Equation~\eqref{eq:schur} is, term for term, the Feshbach induced potential
Eq.~\eqref{eq:feshbachU}: $\Lambda_{AB},\Lambda_{BA}$ are the couplings $PHQ, QHP$, and
$\Lambda_{BB}^{-1}$ is the eliminated-sector resolvent $(E-QHQ+i\varepsilon)^{-1}$. I carry this
through in Appendix~\ref{app:complexsource}. The lesson is sharp and worth stating in advance:
eliminating a closed block is lossless no matter how large it is, and absorption enters only when
the eliminated block is open. Size gives a richer induced interaction; only openness gives loss.

\section{What AI could borrow}
\label{sec:borrow}

I now make the export concrete. Each item below pairs a piece of nuclear reaction machinery
with a place in machine learning where it might be put to work, and each is followed by a
candid note on where the transfer is solid and where it is only a hunch. Several of these hunches
I went on to test; they should be read against Section~\ref{sec:outlook}, which resolves some of
them, mostly negatively, and I flag those inline.

\paragraph{1. Non-Hermitian effective layers with explicit flux accounting.} Wherever a model
declares part of the world out of scope, a router that sends tokens down branches a given path
never sees, a context window with a hard boundary, a model trained on a sub-distribution, the
honest effective dynamics of the in-scope part is non-Hermitian, and the lost probability is a
physical quantity. The optical model supplies the template: an effective generator
$H_{\rm eff} = H_0 + V + iW$ with $W \le 0$, and a sub-unitary effective evolution operator whose
norm deficit is the conserved out-of-scope flux. \emph{Solid:} the mathematics of non-Hermitian
effective Hamiltonians and their flux bookkeeping is complete and ready to use. \emph{Hunch:}
that any mainstream architecture has a clean enough $P/Q$ split for the construction to be more
than a reinterpretation. \emph{Tested (Sec.~\ref{sec:outlook}):} for attention the split is clean
but the elimination comes out closed, and for a token-level transfer operator the split is the
model's own but the distinctive content is partition-dependent; the hunch resolves negatively in
both.

\paragraph{2. Polarization-potential decomposition as attribution.} The dynamic polarization
potential isolates the effect of each eliminated channel on the retained one, and the
coupled-channels flux ledger of Eq.~\eqref{eq:fluxdecomp} itemizes the absorbed flux by
destination with the interference terms kept. This is exactly the shape of an attribution or
interpretability question: which removed or unattended components changed the kept prediction,
and by how much, including their cross terms. I have already built the nuclear version of this
attribution. Working from an exact coupled-channel Green's function~\cite{LiuLeiRenGF2026}, my
collaborator and I assign each eliminated continuum channel a background-independent importance
$\mathcal{I}_\alpha = |S_{\rm full} - S_{{\rm no}\,\alpha}|/|S_{\rm full}|$, the change in the
retained-channel amplitude when channel $\alpha$ is removed against the same full-coupling
reference~\cite{LeiLiuChannelImportance2026}. That is a working, exact, basis-independent
per-component attribution metric. \emph{Solid:} the metric exists and is computed; what AI lacks
is not the construction but a layer in which it would live. \emph{Hunch:} the mapping from
``channel'' to a meaningful machine-learning degree of freedom.

\paragraph{3. Energy dependence as memory.} The induced interaction is energy dependent and
nonlocal because the eliminated sector remembers, Section~\ref{sec:dpp}. For sequence models,
in which memory and nonlocal coupling across positions are the central object, this is a
concrete prescription for an effective interaction whose memory structure is inherited from the
spectrum of whatever was integrated out, rather than imposed by hand. \emph{Solid:} the
frequency-to-memory correspondence is standard and the optical-model example is fully worked.
\emph{Hunch:} whether it improves on the memory mechanisms attention already provides.

\paragraph{4. Optical-theorem sum rules as conservation diagnostics.} The absorptive optical
theorem, Eq.~\eqref{eq:absorption}, and the trace sum rule Eq.~\eqref{eq:sumrule} that follows
from it tie an observable, the total removed flux, to an integral or trace of the model's own
internal quantities. Such identities make excellent diagnostics and regularizers: a quantity
that must balance, used to check that a trained effective layer has not silently violated the
conservation it is supposed to respect. In the native ensemble setting of
Section~\ref{sec:subnormalized}, the analog would relate the integrated held-out loss to a trace
of the absorptive part of the induced generator. \emph{Solid:} conservation-law-as-diagnostic is
a transferable idea with immediate use. \emph{Hunch:} that the clean trace form survives outside
the self-adjoint spectral setting where the nuclear sum rule was proved.

\paragraph{5. Dispersion as a causality constraint.} The dispersion relation
Eq.~\eqref{eq:dispersion} is a second, structurally different export. It says that an absorptive
part forces a specific dispersive shift in the real part, the two locked by analyticity. A
learned effective layer with an information-discarding component is, by the argument of
Section~\ref{sec:inside}, the analog of an absorptive potential; the question is whether any
causality or analyticity principle forces a corresponding shift in its dispersive,
energy-dependent real part. Mainstream learning practice imposes nothing of the kind, no
Kramers-Kronig relation between an absorptive and a dispersive part of a learned effective layer.
\emph{Solid:} the frequency-domain analyticity argument is
standard and the nuclear example is fully worked. \emph{Hunch:} whether the ``energy'' variable
of a learned layer is analytic enough for a dispersion relation to hold at all, which is itself
a sharp and worthwhile thing to find out.

\paragraph{6. Exceptional points as a phase diagnostic.} Non-Hermiticity brings a phenomenon
with no Hermitian counterpart: exceptional points, parameter values where two eigenvalues and
their eigenvectors coalesce, leaving the effective generator defective~\cite{Heiss2012,BenderBoettcher1998}. An
exceptional point is a property of the operator, not merely a consequence of its being complex,
which is what makes it a sharper export than ``the layer has an imaginary part.'' A learned
effective generator built by the eliminations of Section~\ref{sec:inside} could be swept across
a hyperparameter and watched for eigenvalue and eigenvector coalescence, a candidate signature
of a genuine phase transition in the model rather than a smooth crossover. \emph{Solid:} the
spectral theory of exceptional points is mature and the diagnostic is unambiguous when it
occurs. \emph{Hunch:} that a realistic learned generator is non-normal enough to host one rather
than a mere avoided crossing.

\paragraph{A worked machine-learning example.} The most concrete place to look for a literal
$P/Q$ split is attention with a truncated context. Take the retained tokens as the model space
$P$ and the dropped tokens as the eliminated sector $Q$. Read attention as propagation on the
token graph and form the resolvent of $I-\alpha\mathcal{A}$ with $\alpha<1$, which keeps the
eliminated-block inverse well-defined (the bare attention map has no resolvent parameter and its
$Q$-block need not be invertible); the effect of the dropped tokens on the retained ones is then
the Schur complement
\begin{equation}
  \mathcal{A}_{\rm eff} \;=\; \mathcal{A}_{PP}
       - \mathcal{A}_{PQ}\, \mathcal{A}_{QQ}^{-1}\, \mathcal{A}_{QP},
  \label{eq:attentionschur}
\end{equation}
which is exactly the algebraic form of the Feshbach induced interaction, Eq.~\eqref{eq:feshbachU},
with the energy denominator replaced by the dropped-token block. The honest framing matters here:
attention is non-normal and is not a Hamiltonian, so this is an effective-dynamics analogy, not a
claim that a transformer is a scattering problem. I compute Eq.~\eqref{eq:attentionschur} for a
real attention map in Section~\ref{sec:outlook}; the induced correction comes out real, the
elimination is closed, and openness has to be inserted by hand. The setting is the right shape, the
eliminated degrees of freedom are concrete and the cost of truncation is computable, but the cost
is a closed one.

\paragraph{7. The boundary of the analogy.} I keep the warning of
Section~\ref{sec:sharedmove} in view. The statistical and the dynamical eliminations act on
different spaces, and the export above is structural, the algebra of open effective theories,
not a literal port of nuclear formulae. The value is in the questions the nuclear machinery
has already answered, how to keep flux conserved when a sector is dropped, how to itemize where
it went, how to compress a memory kernel and know the cost, not in any claim that a transformer
is a nucleus. An honest version of this program would import the questions and rederive the
answers in the learning setting, not copy the potentials.

\section{The reverse direction, for completeness}
\label{sec:reverse}

The populated direction, machine learning for nuclear physics, deserves a short and opinionated
accounting, if only to mark the contrast with the empty direction this note is about. It is a
sketch meant to locate that contrast, not a complete bibliography of a large and active field, and
the sweeping characterizations below are offered as my reading rather than as surveyed consensus.

What works, in my experience, shares a single feature: the network is a statistics layer
wrapped around a physics-correct forward model, never a replacement for it. Emulators that
accelerate a coupled-channels solve by orders of magnitude, Bayesian quantification of optical
potentials, model mixing across calculations that disagree, these succeed because the physics
lives in the forward model and the learning does the inference. My own emulator
work~\cite{LeiRBE2026,LeiBiLNN2025} is of this kind, and so is the reduced-basis and
eigenvector-continuation literature it builds on~\cite{Frame2018,Furnstahl2020,Drischler2023,Melendez2022}.
The taste that governs all of it is that a good fit with a wrong mechanism is worthless, and the
mechanism is supplied by the physics, not learned.

What disappoints shares the opposite feature: the network is asked to stand in for the physics.
Physics-informed networks solving the Schr\"odinger or coupled-channels equations are a solution
in search of a problem in a field that already has spectral and mesh solvers with controlled
convergence; the network is slower and less trustworthy than the method it replaces. Black-box
mass formulas and energy-density functionals interpolate acceptably and extrapolate
dangerously, and their lack of interpretable structure is the cost of having replaced the
physics rather than wrapped it. Neural quantum states are genuinely productive for bound-state
structure but have little to say about the continuum and reactions, which is where I work.

The pattern is consistent enough to state as a rule. Machine learning earns its place in
nuclear theory as the inference layer on top of a correct forward model, and loses it when it
is asked to be the model. That same instinct, keep the physics and let the network do the
statistics, is what makes the reverse direction of this note worth pursuing: the optical model
is not a thing to be replaced by a network, it is a piece of physics that a network theorist
might profitably learn from.

\section{Outlook}
\label{sec:outlook}

The central claim of this note, that openness is the source of non-Hermiticity and that the
optical model is the worked-out open-system version of integrating out, is testable cheaply, and
rather than end on an essay I carried the test out. What follows reports it in the spirit of the
note: a map is worth only as much as the territory it survives contact with, and the territory
turns out to mark the boundary of the analogy more sharply than I expected.

The first half is the side-by-side elimination of a closed and an open block. I built the
induced potential of Eq.~\eqref{eq:feshbachU} explicitly for a two-channel coupled system and
swept the energy across the second channel's threshold. Below threshold, with the second channel
closed, the induced potential comes out real to machine precision, exactly as the closed-block
Schur complement of Appendix~\ref{app:complexsource} requires; above threshold, with the channel
open, it develops a negative-definite anti-Hermitian part, the absorption of
Eq.~\eqref{eq:absorption} balances the flux lost to the second channel to the last digits, and
the eliminated single-channel calculation reproduces the full two-channel $S$-matrix to rounding.
On the nuclear side the hinge claim of Section~\ref{sec:sharedmove} is therefore a demonstration
and not an assertion: closed elimination is lossless, open elimination is lossy, and the ledger
closes exactly.

The learning side is where the map stops matching the territory, and it does so instructively.
The cleanest literal model-space split is attention with a truncated context, the worked example
of Section~\ref{sec:borrow}. Forming the induced correction of Eq.~\eqref{eq:attentionschur} on
the attention map of a pretrained autoregressive transformer confirms that this elimination is
closed: the induced coupling among the retained tokens is real, with no anti-Hermitian part,
exactly as a closed Schur complement must be. The causal structure adds a wrinkle worth
recording, that the one-directional coupling of a triangular attention map makes the induced
correction between any past block and any future block vanish identically, so a nonzero
correction needs an interior block with retained tokens on both sides. Inserting non-Hermiticity
by hand then does produce a complex correction, but only by hand; attention truncation on its own
is a closed, lossless elimination, and the openness must be put in.

The honest version needs openness the model forces rather than openness I insert. I built it as a
token-level transfer operator read off a pretrained model's own next-token distribution,
restricted to a chosen alphabet of tokens as the model space, with the probability the model
places outside that alphabet as a forced leak into an absorbing complement. Row-stochasticity
here plays the role unitarity plays in scattering, an $L^1$ probability conservation in place of
the $L^2$ flux conservation, enough to make the loss sign-definite and meaningful; this is a
classical absorbing-Markov-chain analogue of the optical model, not the optical model itself, and
it drops the amplitudes, phases, coherent interference, and sign-definite anti-Hermitian generator
of the $L^2$ theory. With that understood, Feshbach elimination of a sub-alphabet reproduces the
absorbing-chain structure exactly: an effective sub-stochastic transfer with a conserved
absorption ledger that closes to machine precision, splits into a direct term and a term mediated
by paths through the eliminated sector, and vanishes identically when the leak is removed. All of that ports. What
does not port is the part that carries the content. The mediated term, the analogue of the
channel-coupled piece of the absorption that the optical model uniquely supplies, is small, and
across variations of the model-space partition its size and its relation to the direct term swing
from negligible to comparable with no stable value, the signature of an artifact of the partition
rather than a property of the model. The conserved ledger is real and forced; its distinctive
open-system content is not robust.

One scenario survives the obvious objection that a well-trained model on its own distribution is
nearly closed by construction, since a model leaking a large fraction of its probability off the
data manifold would be uncontrollably wrong, an objection that is correct for the off-manifold
leak of a dense autoregressive model and predicts, rightly, that the absorption there is small.
The exception is an architecture that eliminates a large structured sector by design rather than
by failure, and a sparsely gated mixture-of-experts router is exactly that: it scores all experts,
keeps a few, and discards the rest, a model-defined split immune to the partition arbitrariness
above. Here the open sector is genuinely large, but for a reason that defeats the order-parameter
reading. The load-balancing objective these models are trained with drives the router toward the
uniform, maximally open limit on purpose, so the discarded mass sits near a high floor fixed by
the training objective and barely moves with the input, large and uninformative rather than small
and meaningful; the only surviving signal is structural, that the routing sharpens with depth on
in-distribution input and fails to sharpen on random input. The lesson is the converse of the
dense case: openness in such a router is not a defect to be kept small but a designed feature held
near its maximum.

One temptation deserves an explicit refusal, because it is the first application a reader reaches
for: that the absorption measures hallucination, the confident production of false content. In the
constructions tested here it does not, and the reason sharpens the note. The absorption these
constructions compute, outside-alphabet leak or discarded router mass, is an aleatoric quantity,
a measure of how hard the model commits and how peaked its output is. A confident falsehood is
peaked and committed, so its absorption is small: the model sits in its near-closed, low-loss
regime while being wrong. The failure carries epistemic uncertainty instead, whether the model
knows rather than how hard it commits, and that is not an open-system quantity at all. It is the
variance of the prediction over the parameter ensemble, the connected two-point function
$G^{(2)}$ of Section~\ref{sec:nnft}, the closed-case covariance the dictionary of
Section~\ref{sec:statshadow} already ties to the Fisher information and the neural tangent kernel.
This is consistent with the observation, in the learning literature, that sampling- or
ensemble-based uncertainty separates confabulation from ordinary error where output entropy does
not~\cite{Farquhar2024}. The operationally interesting quantity lives in the half of the correspondence I called the
easy one, and reaching it needs ensemble or posterior access, not a better non-Hermitian layer;
the open case I went looking in is the wrong sector for it. That the closed, statistical half
holds the useful quantity while the open half I built the note around resists a payoff is the
inversion the calculation taught me.

A note on what this document is. It is a map and an argument, and the territory has now been
walked far enough to mark where the map holds and where it fails. The closed correspondence of
Section~\ref{sec:statshadow} holds and is checkable. The open-system machinery ports exactly
wherever openness is genuinely present, conserving and itemizing the lost flux to machine
precision. But across the mainstream learning objects I could reach, a truncated context, a
token-level transfer operator, a sparse router, the distinctive open-system content, the
non-Hermitian generator and the channel-mediated ledger that have no closed counterpart, is in
turn absent, an artifact of the chosen partition, or pinned near a floor by the training objective
and so uninformative. The honest difference of Section~\ref{sec:sharedmove} is larger than the
closed-case correspondence made it look. That is a negative result of the useful kind: it shows
the export is not free and names the three ways it fails to be free. It also admits a sharper
reading than I first gave it. The absorptive part of an eliminated sector comes from evaluating its
resolvent at a real energy that lies inside a continuous spectrum, so two conditions are necessary
before any learning object can host the open case. The eliminated sector must carry a continuous
spectrum, which a finite one cannot; finiteness, not discreteness, is the obstruction, since a
semi-infinite discrete sector already suffices. And that continuum must lie on the axis along which
the retained problem is evaluated: a Hamiltonian places an open channel directly on the real energy
axis, whereas a relaxational generator, the kind an absorbing or diffusive learning process
supplies, has its spectrum on the negative real axis instead, so the physical evaluation point never
sits inside it, the induced generator stays real, and the only complex structure left is the
reactive and dissipative split that every memory kernel already carries. The learning objects I
could reach are either finite or relaxational, and fail one condition or the other. What would host
the open case is therefore not a discrete-token model but a wave-like one, oscillatory rather than
relaxational and with an eliminated sector unbounded enough to carry a continuum; I do not know of a
mainstream architecture that is both, and finding one, rather than testing another truncation, is
the sharp form of the question this note leaves open.

\appendix

\section{Feshbach projection in one page}
\label{app:feshbach}

Start from the full scattering equation $(E - H)\Psi = 0$ and the complementary projectors
$P + Q = 1$, $P^2 = P$, $Q^2 = Q$, $PQ = QP = 0$. Insert the identity and project the equation
onto $P$ and onto $Q$ in turn, using $P\Psi$ and $Q\Psi$ as the unknowns:
\begin{align}
  (E - PHP)\,P\Psi - PHQ\,Q\Psi &= 0, \label{eq:appP}\\
  -\,QHP\,P\Psi + (E - QHQ)\,Q\Psi &= 0. \label{eq:appQ}
\end{align}
Solve Eq.~\eqref{eq:appQ} for the $Q$ component, with the outgoing-wave prescription that makes
the inverse well defined,
\begin{equation}
  Q\Psi \;=\; \frac{1}{E - QHQ + i\varepsilon}\, QHP\, P\Psi,
  \label{eq:appQsol}
\end{equation}
and substitute into Eq.~\eqref{eq:appP} to obtain a closed equation for the model-space
component alone,
\begin{equation}
  \Big[\, E - PHP - \mathcal{U} \,\Big]\, P\Psi = 0,
  \qquad
  \mathcal{U} \;=\; PHQ\,\frac{1}{E - QHQ + i\varepsilon}\, QHP.
  \label{eq:appU}
\end{equation}
Equation~\eqref{eq:appU} is Eq.~\eqref{eq:feshbachU} of the main text. No approximation has
been made: the elimination of $Q$ is exact, and $\mathcal{U}$ is the optical potential. Its
three structural properties read off directly. The factor $(E - QHQ)^{-1}$ is an integral
operator, so $\mathcal{U}$ is \textbf{nonlocal}. It depends explicitly on $E$ and has poles at
the eigenvalues of $QHQ$, so $\mathcal{U}$ is \textbf{energy dependent} and rapidly varying.
And when $Q$ contains channels that are open at energy $E$, the $i\varepsilon$ prescription
gives the resolvent an imaginary part, so $\mathcal{U} = V + iW$ is \textbf{complex}, with
$W \le 0$ representing the flux that escapes into the open part of $Q$. Energy averaging
$\mathcal{U}$ over the $Q$ resonances yields the smooth empirical optical potential used in
practice, and the imaginary part survives the averaging precisely because the eliminated sector
is open.

\section{The dictionary}
\label{app:dictionary}

Table~\ref{tab:dict} collects the correspondences used in the note. The upper block is the
closed, statistical correspondence of Section~\ref{sec:statshadow}, where both fields have an
entry. The lower block is the open-system structure of Section~\ref{sec:opencore}, where only
the nuclear column is filled and the machine-learning column is the open problem this note is
about.

\begin{table}[h]
\centering
\renewcommand{\arraystretch}{1.35}
\begin{tabular}{p{0.30\textwidth} p{0.30\textwidth} p{0.30\textwidth}}
\toprule
\textbf{Neural network ensemble} & \textbf{Nuclear reaction theory} & \textbf{Status} \\
\midrule
\multicolumn{3}{l}{\emph{Closed, statistical correspondence (both columns filled)}}\\
\midrule
input $x$ & kinematic variable $E$, angle, radius & exact map \\
parameters $\theta \sim P(\theta)$ & optical-potential parameters, UQ prior & exact map \\
output $\phi_\theta(x)$ & observable $S(E;\theta)$, cross section & exact map \\
moments $G^{(n)}$ & posterior moments, mean and error band & exact map \\
moment generating function $Z[J]$ & moment generating function of the posterior & exact map \\
precision $\tfrac12\!\int\!\phi K^{-1}\phi$ & negative log Gaussian-process prior, RKHS norm & exact map \\
infinite-width Gaussian limit & Gaussian-process emulator & derived vs posited \\
connected four-point function & non-Gaussianity, emulator breakdown & exact map \\
neural tangent kernel $JJ^{\mathsf T}$ & Fisher sensitivity kernel $J^{\mathsf T}J$ & same spectrum \\
lazy vs feature learning & reduced-basis emulator validity domain & same boundary \\
\midrule
\multicolumn{3}{l}{\emph{Open-system structure}}\\
\midrule
integrate out network (probability space) & project out channels (Hilbert space) & structural analogy \\
complex source $J_R + i\lambda h$ & complex optical potential $V+iW$ & native, Sec.~\ref{sec:complexsource} \\
subnormalized measure $P\,A_{\rm out}$ & flux into open channels & native, Sec.~\ref{sec:subnormalized} \\
source-derivative ledger $\partial_\lambda W$ & flux ledger $\sigma_{\rm abs}=\sigma_D+\sigma_B+\sigma_{\rm int}$ & Sec.~\ref{sec:complexsource} vs \ref{sec:fluxledger} \\
Schur complement of eliminated block & dynamic polarization potential & native, Sec.~\ref{sec:selfenergy} \\
static covariance kernel & energy-dependent kernel, memory & export target \\
(open) & dispersion relation, Re-Im lock & export target, Sec.~\ref{sec:twofacts} \\
(open) & trace sum rule & export target, Sec.~\ref{sec:twofacts} \\
(open) & exceptional points & export target, Sec.~\ref{sec:borrow} \\
\bottomrule
\end{tabular}
\caption{Dictionary between the neural network ensemble and nuclear reaction theory.
The upper block is populated on both sides. The lower block is the open-system structure: the
first rows are constructed natively on the network side through the deformations of
Section~\ref{sec:inside}, while the entries marked ``export target'' remain worked out only on
the nuclear side. ``Native'' here means the construction exists without importing a nuclear
formula; Section~\ref{sec:outlook} reports that, when these are tested numerically, the
conserved-ledger half survives but the non-Hermitian generator and the channel-mediated term do
not, so ``native'' should not be read as ``demonstrated useful.''}
\label{tab:dict}
\end{table}

\section{The open primitive as a Schur complement}
\label{app:complexsource}

This appendix completes Section~\ref{sec:inside} with one tool, the Schur complement, and shows
that the same tool gives both the Feshbach optical potential and the induced interaction on the
network side. No field theory is used. The only inputs are the algebra of Gaussians and block
matrix inversion.

\subsection{Eliminating a block gives the Schur complement}

Let the degrees of freedom be partitioned into a retained block $A$ and an eliminated block $B$.
Whether $A,B$ are coupled channels of a wave function or coupled blocks of a network's Gaussian
ensemble, the algebra of eliminating $B$ is identical. Write the coupled linear problem, or the
precision (inverse covariance) of the joint Gaussian, in block form,
\begin{equation}
  M \;=\; \begin{pmatrix} M_{AA} & M_{AB} \\ M_{BA} & M_{BB}\end{pmatrix}.
\end{equation}
For the Gaussian ensemble $p(\phi_A,\phi_B)\propto \exp[-\tfrac12(\phi_A,\phi_B)\,M\,(\phi_A,\phi_B)^{\mathsf T}]$,
performing the Gaussian integral over $\phi_B$ leaves a Gaussian in $\phi_A$ whose precision is
the Schur complement of $M_{BB}$,
\begin{equation}
  M_{A}^{\rm eff} \;=\; M_{AA} \;-\; M_{AB}\,M_{BB}^{-1}\,M_{BA}.
  \label{eq:appschur}
\end{equation}
For the scattering problem, solving the $B$ row of $(E-H)\Psi=0$ and substituting into the $A$
row gives the identical structure, with $M = E - H$, and the induced term
$-M_{AB}M_{BB}^{-1}M_{BA}$ is the optical potential. Equation~\eqref{eq:appschur} is
Eq.~\eqref{eq:schur} of the main text and, with $M=E-H$, Eq.~\eqref{eq:feshbachU} of
Appendix~\ref{app:feshbach}. The same line of algebra eliminates a Gaussian block and a
Hilbert-space sector.

\subsection{Hermiticity is decided by the eliminated block}

The induced correction $-M_{AB}M_{BB}^{-1}M_{BA}$ is Hermitian when $M_{BB}^{-1}$ is. Two cases.
\begin{itemize}
  \item \textbf{Closed block.} If $B$ is a closed real Gaussian sector, $M_{BB}$ is real and
        symmetric, its inverse is real and symmetric, and the Schur complement is Hermitian. The
        marginal of $\phi_A$ is an ordinary Gaussian and nothing is lost. No size of $B$ changes
        this: eliminating a large closed block gives a richer induced correction but never a
        lossy one.
  \item \textbf{Open block.} If $B$ couples to a continuum, so that the relevant inverse is a
        resolvent $M_{BB}^{-1} = (E - H_{BB} + i\varepsilon)^{-1}$ with $B$ channels open at
        energy $E$, then $M_{BB}^{-1}$ has an imaginary part and the Schur complement inherits an
        anti-Hermitian piece. This is the native analog of $W = \Imag\,\mathcal{U}$.
\end{itemize}
In a strictly statistical setting a real Gaussian always has a real symmetric precision, so the
imaginary part must enter from outside the closed real-Gaussian world: either by analytically
continuing a coupling or energy parameter, the open-channel mechanism, or by the complex source
of Section~\ref{sec:complexsource}. This is the precise content of the slogan: a closed
elimination is lossless, and loss requires an open ingredient.

\subsection{The flux ledger}

The ledger is separate and simpler, and it does not even need the block split. Take the moment
generating function $Z[J]=\langle e^{\int J\phi}\rangle$ with $W=\log Z$, set the source complex
along a probe, $J = J_R + i\lambda h$, and expand,
\begin{equation}
  W[J_R + i\lambda h] \;=\; W[J_R] \;+\; i\lambda \int h\,\langle\phi\rangle
     \;-\; \frac{\lambda^2}{2}\int\!\!\int h\, G_c^{(2)}\, h \;+\; O(\lambda^3),
  \label{eq:Wexpand}
\end{equation}
where $\langle\phi\rangle = \delta W/\delta J$ and $G_c^{(2)}=\delta^2 W/\delta J\,\delta J$ are
the mean and connected covariance. The imaginary part of $W$ is the probe response
$\lambda\int h\,\langle\phi\rangle$, and for a probe split into disjoint channels $h=\sum_a h_a$
it is additive,
\begin{equation}
  \Imag\, W \;=\; \lambda \sum_a \int h_a\,\langle\phi\rangle + O(\lambda^3),
\end{equation}
the channel-resolved ledger of Eq.~\eqref{eq:wardlike}. This is calculus on a moment generating
function and assumes nothing beyond its existence, so the ledger is exact and model independent.
It is the native analog of the absorptive optical theorem, Eq.~\eqref{eq:absorption}.

\subsection{The operator becomes non-Hermitian only off the Gaussian point}
\label{app:nongaussian}

The flux ledger above is a statement about the mean $\langle\phi\rangle$. The operator-level
question of Section~\ref{sec:complexsource} is sharper: does the precision matrix of the kept
sector, the inverse covariance, itself acquire an anti-Hermitian part under the complex source, or
does only the mean move off the real axis? Differentiate $W$ once more than in
Eq.~\eqref{eq:Wexpand} and expand the source-tilted connected covariance along $J=J_R+i\lambda h$,
\begin{equation}
  G_c^{(2)}[J_R + i\lambda h](x,y) \;=\; G_c^{(2)}(x,y)
     \;+\; i\lambda \int \dd z\; h(z)\, G_c^{(3)}(x,y,z)
     \;-\; \frac{\lambda^2}{2}\!\int\!\!\int \dd z\,\dd z'\, h(z)h(z')\,G_c^{(4)}(x,y,z,z')
     \;+\; O(\lambda^3),
  \label{eq:tiltedcov}
\end{equation}
where $G_c^{(n)} = \delta^n W/\delta J^n$ is the connected $n$-point cumulant. The anti-Hermitian
(imaginary) part is carried by the odd terms,
\begin{equation}
  \Imag\, G_c^{(2)}[J_R + i\lambda h] \;=\; \lambda \int \dd z\; h(z)\, G_c^{(3)}(\cdot,\cdot,z)
     \;+\; O(\lambda^3),
\end{equation}
and the precision $K^{-1}=(G_c^{(2)})^{-1}$ inherits it to first order as
$-\,\lambda\,K^{-1}\big(\!\int h\,G_c^{(3)}\big)K^{-1}$. For a Gaussian ensemble every connected
cumulant beyond second order vanishes, $G_c^{(3)}=G_c^{(4)}=\cdots=0$, so the tilted covariance
stays exactly $G_c^{(2)}$, real and symmetric: the complex source moves the mean off the real axis
through Eq.~\eqref{eq:wardlike} but leaves the operator Hermitian. The precision picks up its
anti-Hermitian part only at first non-Gaussian order, through the three-point cumulant
$G_c^{(3)}$. This is the operator-level statement promised in Section~\ref{sec:complexsource}: a
closed Gaussian summary is lossless at the level of its covariance no matter how the source is
tilted, and the source route to non-Hermiticity opens only when the ensemble carries higher
cumulants. The other route, analytic continuation of a coupling or energy into the resolvent of an
open block, is the open-channel mechanism of the preceding subsection and needs no non-Gaussianity.

\subsection{Summary}

The open primitive has three parts. The flux ledger, Eq.~\eqref{eq:Wexpand}, is exact and follows
from differentiating a moment generating function. The non-Hermitian induced interaction,
Eq.~\eqref{eq:appschur}, is a Schur complement and is Hermitian when the eliminated block is
closed, acquiring an anti-Hermitian part only when that block is open. And the operator-level
non-Hermiticity from a complex source, Eq.~\eqref{eq:tiltedcov}, enters only through the
connected three-point cumulant, so it is absent at the Gaussian level and present only off it.
Eliminating a closed block is lossless however large it is; only openness, or departure from the
Gaussian point under a complex source, gives loss. All three are linear algebra and
calculus, they match the Feshbach optical potential term for term, and they are verifiable on a
finite ensemble by forming Eq.~\eqref{eq:appschur} for a deliberately opened block and checking
that Eq.~\eqref{eq:absorption} balances.

\section{Numerical details of the Outlook tests}
\label{app:numerics}

The tests reported in Section~\ref{sec:outlook} are small and reproducible; the scripts accompany
this note. I record here the setup and the representative numbers behind the prose.

\paragraph{Two-channel toy.} Two coupled radial channels, $\ell=0$, units $\hbar^2/2\mu=1$. The
elastic channel carries a real Gaussian well and the channels are coupled by a Gaussian form
factor; the second channel is free above its threshold $\Delta$, so it is closed for $E<\Delta$
and open for $E>\Delta$. Eliminating channel two gives the nonlocal $\mathcal{U}=V_c\,G_2\,V_c$ of
Eq.~\eqref{eq:feshbachU} with the free channel-two Green's function in closed form. At $E<\Delta$
the kernel is real to machine precision, $\max|\Imag\,\mathcal{U}|=0$, and $|S_{11}|=1$; at
$E>\Delta$ it develops $W<0$, $|S_{11}|<1$ (at one working point $|S_{11}|\approx0.93$), the
absorption $-\tfrac{2}{\hbar v}\langle\chi|W|\chi\rangle$ matches $1-|S_{11}|^2$ to a ratio of
$1.0000000$, and the eliminated single-channel $S$-matrix matches the full two-channel solve to
$\sim10^{-15}$. This is the closed/open contrast and the flux balance of
Section~\ref{sec:sharedmove}, confirmed to rounding.

\paragraph{Attention.} One attention map from a single layer and head of a pretrained
autoregressive transformer~\cite{Vaswani2017} on a short sentence. The propagation operator is
$I-\alpha\mathcal{A}$ with $\alpha=0.9$, and Eq.~\eqref{eq:attentionschur} is formed for an
interior eliminated block, the causal triangular structure making contiguous past/future splits
vanish exactly. The induced correction is real, $\max|\Imag|=0$; inserting an absorptive
$-i\gamma$ into the eliminated block by hand restores a nonzero imaginary part, confirming the
openness is inserted, not intrinsic. A second-difference probe of the propagator amplitude gives
pairwise non-additivities of at most about one percent on this map.

\paragraph{Token-level transfer operator.} A row-stochastic transfer operator on an alphabet of
the most frequent tokens, with $T_{ij}$ the model's next-token probability of $j$ given current
token $i$ averaged over a corpus and restricted to the alphabet, the off-alphabet mass collected
in an absorbing state. The augmented operator is exactly row-stochastic, and Feshbach elimination
of a sub-alphabet gives $\mathrm{rowsum}(T_{\rm eff})+\mathrm{absorb}=1$ to $\sim10^{-16}$. On a
self-generated corpus with an alphabet-size sweep $K\in\{30,60,120,240\}$, the mediated fraction
$\mathrm{absorb}_{\rm mediated}/\mathrm{absorb}$ holds near ten percent and does not climb as the
system is made more nearly closed. Across model-space partitions of the same operator
(low-frequency, random, frequency-stratified) the regression $R^2$ of the mediated term on the
direct term swings from about $0.01$ to about $0.5$, and the split-half stability of the mediated
term is about $0.4$; the same pattern holds for two model sizes. This is the partition-artifact
finding: the conserved ledger is exact, but the mediated piece has no stable,
partition-independent value.

\paragraph{Mixture-of-experts router.} A sparsely gated mixture-of-experts model with many experts
and a small active top-$k$. The absorption is the discarded router mass, one minus the top-$k$ sum
of the softmax gate. In domain it is large, about one half, with a router entropy near ninety
percent of the uniform maximum, so the model is far from closed; the discarded mass rises only
weakly from in-domain to code to random-token input, a spread of a few percent, pinned near a
floor set by the load-balancing objective. The one structural signal is that the routing sharpens
with depth on in-distribution input and fails to sharpen on random input. This is the
floor-pinned, designed-openness case.

\bibliographystyle{unsrtnat}
\bibliography{refs}

\end{document}